\documentclass[letterpaper, 10 pt, journal, twoside]{ieeetran}

\IEEEoverridecommandlockouts                              

\usepackage{graphics} 
\usepackage{epsfig} 
\usepackage{mathptmx} 
\usepackage{times} 
\usepackage{amsmath} 
\usepackage{lipsum}
\usepackage{mathtools}
\usepackage{cuted}
\usepackage{arydshln}
\usepackage{subcaption}
\usepackage{hyperref}

\usepackage{color}
\usepackage{booktabs}
\usepackage{array}
\usepackage{tabularx}
\usepackage{caption}
\usepackage{cite}
\usepackage{newtxtext,newtxmath}

\definecolor{somegray}{rgb}{0.5, 0.5, 0.5}
\newcommand{\darkgrayed}[1]{\textcolor{somegray}{#1}}
\makeatletter
\newcommand*\titleheader[1]{\gdef\@titleheader{#1}}
\AtBeginDocument{%
  \let\st@red@title\@title
  \def\@title{%
    \vskip-2em
    \bgroup\normalfont\large\centering\@titleheader\par\egroup
    \vskip1.5em\st@red@title}
}
\makeatother

\titleheader{\darkgrayed{This paper has been accepted for publication at the\\
IEEE Robotics and Automation Letters, 2021.
\copyright IEEE}}

\title{Autonomous Quadrotor Flight despite Rotor Failure\\ with Onboard Vision Sensors: Frames vs. Events }

\begin{document}

\author{Sihao Sun$^1$, Giovanni Cioffi$^1$, Coen de Visser$^2$, Davide Scaramuzza$^1$
\thanks{$^1$S.Sun, G. Cioffi and D. Scaramuzza are with the Robotics and Perception Group, University of Zurich, Switzerland (\protect\url{http://rpg.ifi.uzh.ch}).} 
\thanks{$^2$C. de Visser is with Faculty of Aerospace Engineering, Delft University of Technology, 2629 HS Delft, The Netherlands.}
}
\maketitle

\begin{abstract}
Fault-tolerant control is crucial for safety-critical systems, such as quadrotors. State-of-art flight controllers can stabilize and control a quadrotor even when subjected to the complete loss of a rotor. However, these methods rely on external sensors, such as GPS or motion capture systems, for state estimation. To the best of our knowledge, this has not yet been achieved with only onboard sensors. In this paper, we propose the first algorithm that combines fault-tolerant control and onboard vision-based state estimation to achieve position control of a quadrotor subjected to complete failure of one rotor. 
Experimental validations show that our approach is able to accurately control the position of a quadrotor during a motor failure scenario, without the aid of any external sensors. The primary challenge to vision-based state estimation stems from the inevitable high-speed yaw rotation (over 20 rad/s) of the damaged quadrotor, causing motion blur to cameras, which is detrimental to visual inertial odometry (VIO). We compare two types of visual inputs to the vision-based state estimation algorithm: standard frames and events. Experimental results show the advantage of using an event camera especially in low light environments due to its inherent high dynamic range and high temporal resolution. We believe that our approach will render autonomous quadrotors safer in both GPS denied or degraded environments. We release both our controller and VIO algorithm open source.


\end{abstract}

\section*{Source code and video}
The source code of both our controller and VIO algorithm is available at:
\url{https://github.com/uzh-rpg/fault_tolerant_control}

A video of the experiments is available at:
\url{https://youtu.be/Ww8u0KH7Ugs}

\section{Introduction}\label{sec:introduction}

\IEEEPARstart{M}{ulti-rotor} drones have been widely used in many applications, such as inspection, delivery, surveillance, agriculture, and entertainment. Among different types of multi-rotor drones, quadrotors are the most popular by virtue of their simple structure and relatively high aerodynamic efficiency. However, due to less rotor redundancy, quadrotors are also more vulnerable to motor failures. 

Because safety is always a major concern, which restricts the expansion of the drone industry, it is crucial to develop methods to prevent quadrotors from crashing after motor failures.
Fault-tolerant flight control of quadrotors is a feasible solution since only software adaptation is required, which is an obvious advantage over adding rotor redundancy or parachutes. 

Previous works have achieved autonomous flight of a quadrotor subjected to complete failure of a single rotor \cite{Mueller2015,Stephan2018}, and even in high-speed flight conditions where aerodynamic disturbances are apparent \cite{Sun2018,Sun2020}. However, these methods rely on external sensors, such as GPS or motion capture systems, for position tracking, which completely eliminate or alleviate the effects of state estimation errors. To improve quadrotor safety in GPS denied or degraded environments, we need to resort to fully onboard solutions, such as vision-based state estimation.

\begin{figure}[t!]
\begin{center}
\includegraphics[width=1.0\linewidth]{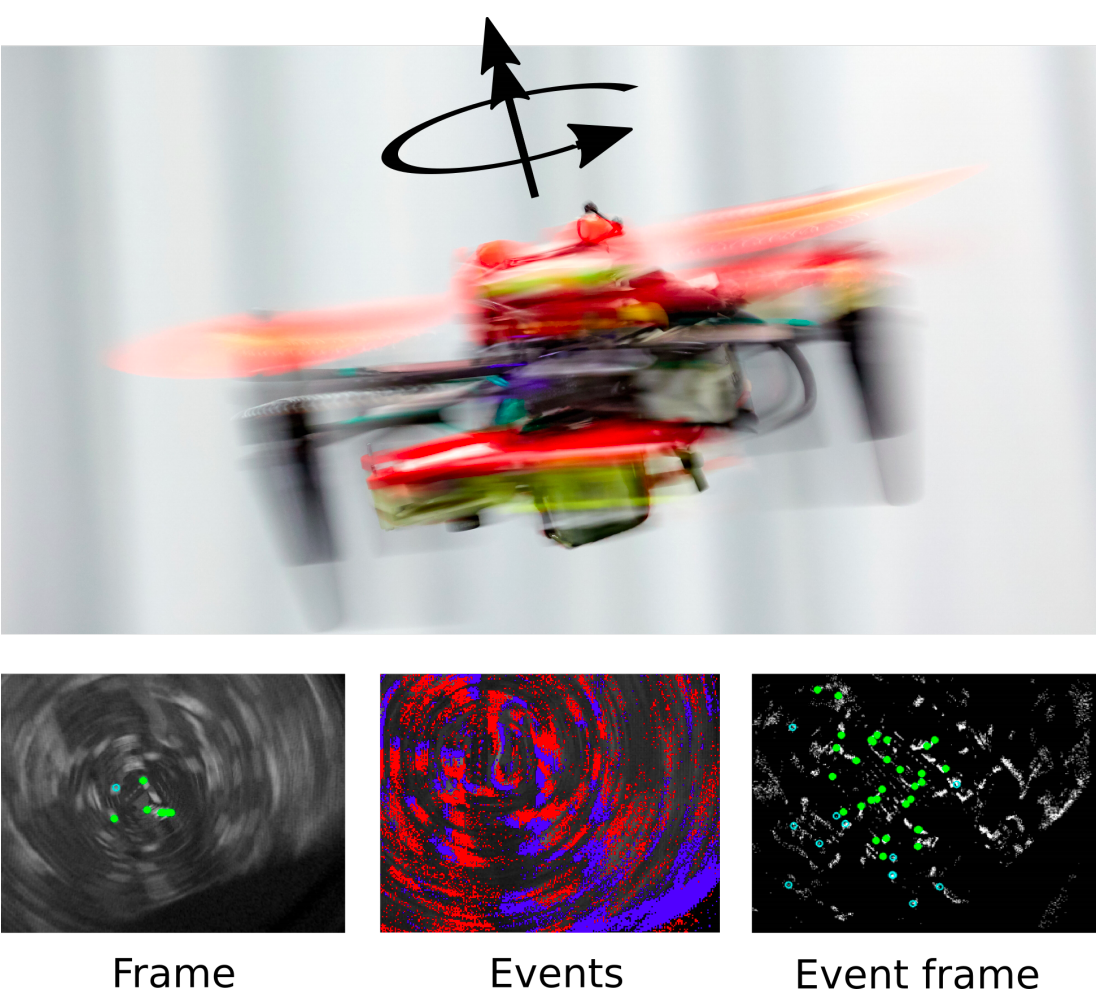}
\end{center}
  \caption{Controlling a quadrotor with complete failure of a rotor causes fast yaw rotation, over 20 rad/s (top figure). Bottom figures show a standard frame and events captured by an onboard event camera. \textbf{Bottom Left}: standard frame with motion blur. \textbf{Bottom Center}: Events only (blue: positive events, red: negative events). \textbf{Bottom Right}: Event frame generated from events. Blue circles are detected features, green dots indicate tracked features.}
\label{fig:intro_figure}
\end{figure}

Complete failure of a rotor results in a fast spinning motion ($>20$ rad/s) of the quadrotor \cite{freddi2011feedback}. This high-speed motion brings a significant challenge to onboard state-estimation methods. First, in vision-based estimators, it causes motion blur (bottom left plot in Fig.~\ref{fig:intro_figure}). Such motion blur deteriorates feature detection and tracking and subsequently degrades  visual-inertial odometry (VIO) performance, especially in low-light environments. Secondly, a large centrifugal acceleration read by the inertial measurement unit (IMU) often causes large errors in commonly-used attitude estimators. These problems need to be resolved to achieve autonomous quadrotor flight despite rotor failure, using only onboard sensors.

\subsection{Contributions}
To address the aforementioned problems, we make the following contributions:

\begin{enumerate}

    \item We achieve and demonstrate the first closed-loop flight of a quadrotor subjected to complete failure of a rotor, using only onboard sensors and computation. We demonstrate that a damaged quadrotor can both hover as well as follow a sequence of setpoints.  
    
    \item We propose a novel state-estimation algorithm combining measurements from an IMU, a range sensor, and a vision sensor that can be either a standard camera or an event camera~\cite{gallego2020event}. We show that an event camera becomes advantageous in low-light environments because the sensor does not suffer from motion blur and has a very high dynamic range.


    \item We improve the complementary filter, commonly used for attitude estimation, to account for the measurement error (up to $1g$) induced by the fast spinning motion on the accelerometer. We show that this yields significant improvements in pitch and roll estimates in this high-speed spinning flight condition.

\end{enumerate}

\subsection{Related Work}
\subsubsection{Quadrotor Fault-Tolerant control}
The first flight controller resilience to complete failure of a single rotor was proposed by \cite{freddi2011feedback} using a feedback linearization approach. The authors demonstrated that sacrificing the stability in the yaw direction becomes inevitable in order to keep the full control of pitch, roll, and thrust. As a consequence, the drone fast spins due to the imbalanced drag torques from remaining rotors. Following this idea, different approaches were proposed for this problem, such as PID~\cite{Lippiello2014}, backstepping~\cite{Lippiello2014a}, robust feedback linearization \cite{Lanzon2014}, and incremental nonlinear dynamic inversion (INDI)~\cite{Lu2015}. However, these works were only validated in simulation environments.

The first real-world flight test of a quadrotor with a complete failure of a rotor was achieved by~\cite{Mueller2014}, where a linear quadratic regulator (LQR) was applied. The authors also proposed the \textit{relaxed-hovering solution} as a quasi-equilibrium, where the quadrotor spins about a fixed point in the 3-D space, with constant body rates, pitch, and roll angles~\cite{Mueller2015}. 
The authors of \cite{Stephan2018} considered the initial spinning phase with varying yaw rate, using a linear parameter varying controller. As the above controllers did not consider aerodynamic effects, the INDI approach was applied in \cite{Sun2018} and \cite{Sun2020} to render the controller resilient to significant aerodynamic disturbances in high-speed wind-tunnel flight tests.

\subsubsection{Visual Inertial Odometry}
VIO~\cite{huang2019icra, zhang2019encyclopedia} fuses visual and inertial measurements to estimate the 6-DoF pose of the sensor rig.
Recent progress has made VIO algorithms computationally efficient to be deployed on resource constrained platforms such as quadrotors~\cite{delmerico2018icra}.
In this work, we are interested in \textit{sliding-window optimization-based}, also known as \textit{fixed-lag smoothing}, VIO estimators.
Such methods estimate a sliding-window of the most recent system states in a nonlinear least-squares optimization problem.
They are normally more accurate than \text{filter-based} methods~\cite{strasdat2012visual, delmerico2018benchmark}. 
Two successful sliding-window optimization-based estimators are~\cite{leutenegger2015keyframe, qin2018vins}.
These estimators minimize a cost function containing visual, inertial, and marginalization residuals.
A limited number of the most recent system states is kept in the sliding window.
Old states are marginalized together with part of the measurements.
In this work, we use a sliding-window optimization-based approach by virtue of its favorable trade-off between computational efficiency and estimate accuracy.

Recent works~\cite{rebecq2016evo, vidal2018ultimate} showed that event cameras allow to deploy VIO-based state estimators in scenarios where standard cameras are not reliable, such as high dynamic range scenes and high speed motions, which cause large motion blur.
In~\cite{rebecq2016evo}, an event-based algorithm was presented, which is able to estimate the pose of the camera and build a semi-dense 3D map when the camera undergoes fast motion and operates in challenging illumination conditions.
In~\cite{vidal2018ultimate}, both standard and event cameras were combined into a sliding-window-based VIO pipeline.
The authors showed that leveraging the complementary properties of the standard and event cameras leads to large accuracy improvements with respect to the standard frame-only or event-only algorithms.

\subsection{Notation}
Throughout the paper, we use bold lowercase letter to represent vector variables and bold capital letters for matrices; otherwise, they are scalars. Superscript $B$ indicates that the vector is in the body frame, and $C$ stands for the camera frame. By default, a vector without superscript is represented in the inertial frame. A 3-D vector with subscript '$\times$' means its skew-symmetric matrix for cross product, such that $\boldsymbol{a}_\times \boldsymbol{b} = \boldsymbol{a}\times \boldsymbol{b}$ for any $\boldsymbol{a}\in \mathbb{R}^3$. $\mathrm{diag}(a,~b)$ represents diagonal matrix with $a$ and $b$ as diagonal elements.

\subsection{Organization}
This paper is organized as follows: We first introduce the fault-tolerant flight controller in Section \ref{sec:flight_controller_design}. Then Section \ref{sec:state_estimator} details the state estimator, including the VIO algorithm and the improved complementary filter. The evaluation and closed-loop flight results are given in Section \ref{sec:experiments}. Afterwards the conclusions are drawn in Section \ref{sec:conclusions}.
\section{Flight Controller Design}\label{sec:flight_controller_design}
\subsection{Quadrotor Model}
\subsubsection{Quadrotor Dynamic Model}
The translational and rotation dynamic equations of a quadrotor are:

\begin{equation}
    m\ddot{\boldsymbol{\xi}}= m\boldsymbol{g} + \boldsymbol{R}_\mathrm{IB}\boldsymbol{f}^B,
    \label{eq:F_dyn_equation}
\end{equation}
\begin{equation}
    \boldsymbol{I}_v^B\dot{\boldsymbol{\omega}}^B + \boldsymbol{\omega}^B\times \boldsymbol{I}_v^B\boldsymbol{\omega}^B=  \boldsymbol{m}^B + \boldsymbol{m}_g^B,
    \label{eq:M_dyn_equation}
\end{equation}
where $\boldsymbol{\xi} = [x,~y,~z]^T$ indicates the center of gravity (c.g.) location of the drone in the inertial frame. $\boldsymbol{R}_\mathrm{IB} \in$ SO(3) is the rotational matrix representing the quadrotor attitude. The angular velocity of the body frame w.r.t the inertial frame is expressed as $\boldsymbol{\omega}^B = [\omega_x,~\omega_y,~\omega_z]^T$.  The vehicle mass and inertia are denoted by $m$ and $\boldsymbol{I}_v^B$ respectively, and $\boldsymbol{g} = [0,~0,~g]^T$ denotes the gravity vector, where $g$ is the local gravitational constant. $\boldsymbol{m}_g^B$ is the propeller gyroscopic moments. $\boldsymbol{f}^B = [0,~0,~T]^T$ is the external force vector projecting on the body frame.



The collective thrust $T$ and control torques $\boldsymbol{m}^B = [\tau_x,~\tau_y,~\tau_z]^T$ are generated by rotors from the control allocation model:
\begin{equation}
    \left[T,~\tau_x,~\tau_y,~\tau_z\right]^T =
    \boldsymbol{G}\boldsymbol{u},
    \label{eq:control_allocation}
\end{equation}
where $\boldsymbol{u} \in \mathbb{R}^4$ is the control input vector containing thrusts generated by each rotor. $\boldsymbol{G}\in \mathbb{R}^{4\times 4}$ is the \textit{control effective matrix} projecting individual thrust of each rotor to the collective thrust and torques.

\subsubsection{Reduced Control Allocation Model}
After the occurrence of the complete failure of the $i$-th rotor, where $i\in\{1,~2,~3,~4\}$, a quadrotor cannot maintain a stationary condition \cite{freddi2011feedback}. It is a common strategy to stop controlling quadrotor heading angle leading to high-speed yaw rotation. In this case, we remove the last row and the $i$-th column of $\boldsymbol{G}$, and obtain a reduced control effective matrix $\tilde{\boldsymbol{G}} \in \mathbb{R}^{3\times 3}$. The control input $\boldsymbol{u}$ is also reduced to $\tilde{\boldsymbol{u}}\in \mathbb{R}^3$ where the $i$-th element is removed. Then, we obtain the reduced control allocation model
\begin{equation}
    \left[T,~\tau_x,~\tau_y\right]^T =
    \tilde{\boldsymbol{G}}\tilde{\boldsymbol{u}}.
    \label{eq:reduced_control_allocation}
\end{equation}

\subsubsection{Reduced Attitude Kinematic Model}
Since the heading is not being controlled, instead of a full attitude kinematic model, we use a reduced attitude kinematic model~\cite{Sun2018} to design the flight controller:
\begin{equation}
    \dot{\boldsymbol{n}}^B = \boldsymbol{n}^B_{\times}\boldsymbol{\omega}^B + \boldsymbol{R}_\mathrm{IB}^T\dot{\boldsymbol{n}},
    \label{eq:reduced_kinematics}
\end{equation}
where $\boldsymbol{n}$ is an arbitrary unit vector and $\boldsymbol{n}^B=[n_x,~n_y,~n_z]^T$. 

\subsection{Fault Tolerant Controller}
Given $\boldsymbol{a}_\mathrm{des}$ the reference acceleration from a PD position controller \cite{Mueller2014}, we can obtain the desired thrust orientation as $\boldsymbol{n} = (\boldsymbol{a}_\mathrm{des}-\boldsymbol{g})/||(\boldsymbol{a}_\mathrm{des}-\boldsymbol{g})||$. The reduced attitude controller aims at aligning $\boldsymbol{n}$ with a body frame fixed vector $\boldsymbol{n}_\mathrm{fix}^B = [\bar{n}_x, ~ \bar{n}_y, ~ \bar{n}_z]^T$, where $\bar{n}_x, ~ \bar{n}_y, ~ \bar{n}_z$ are constants~\cite{chaturvedi2011rigid}. A straightforward option is to select $\boldsymbol{n}_\mathrm{fix}^B=[0,~0,~1]^T$, namely the thrust direction, to align with $\boldsymbol{n}$. However, we keep a slight angle (around 15~deg) between $\boldsymbol{n}_\mathrm{fix}^B$ and $\boldsymbol{n}$. This reduces the spinning rate and causes wobbling motion of the drone. We refer the reader to \cite{Mueller2014} for more details. 

To align $\boldsymbol{n}_\mathrm{fix}$ with $\boldsymbol{n}$, we use a nonlinear dynamic inversion (NDI) approach~\cite{da2003reentry}. We define the control variable as 
\begin{equation}
\boldsymbol{y} = \left[
    \begin{array}{c}
         n_x - \bar{n}_x  \\
         n_y - \bar{n}_y 
    \end{array}\right]
    =
    \left[
    \begin{array}{c}
         y_1  \\
         y_2 
    \end{array}
    \right].
    \label{eq:y_expression}
\end{equation}
Then we need to design the control input $\tilde{\boldsymbol{u}}$ leading to a stable second-order closed-loop dynamic of $\boldsymbol{y}$:
\begin{equation}
    \ddot{\boldsymbol{y}}(\tilde{\boldsymbol{u}}) = -\mathrm{diag}(k_p,~k_p) \boldsymbol{y} - \mathrm{diag}(k_d,~k_d) \dot{\boldsymbol{y}},
    \label{eq:y_closed_loop}
\end{equation}
where $k_p$ and $k_d$ are proportional and derivative gains. An integral term may be added to (\ref{eq:y_closed_loop}) as well to address slight model mismatch. To simplify the derivation of $\tilde{\boldsymbol{u}}$, we assume that $\omega_z$ and $n_z$ are constant after rotor failure. Similarly to \cite{Mueller2014}, we assume a slowly changing $\boldsymbol{n}$ and $\dot{\boldsymbol{n}}=\boldsymbol{0}$. The quadrotor inertia matrix $\boldsymbol{I}_v^B$ is approximated by $\mathrm{diag}(I_x,~I_y,~I_z)$. Propeller gyroscopic moment $\boldsymbol{m}_g^B$ is also negligible compared with rotor thrust generated torques, thus we omit it in the controller design. We validate these assumptions in the real flights. Then, we substitute (\ref{eq:M_dyn_equation})(\ref{eq:reduced_kinematics})(\ref{eq:y_expression}) into (\ref{eq:y_closed_loop}), yielding expressions of roll and pitch torque commands:

\begin{equation}
\begin{array}{l}
     \left[\tau_{x,\mathrm{cmd}},~\tau_{y,\mathrm{cmd}}\right]^T =  \\
     \\
     \left[
     \begin{array}{c}
         \begin{array}{l}
             \left[k_py_2 + k_d(n_x\omega_z-n_z\omega_x) + (n_y\omega_z-n_z\omega_x)\omega_z\right]I_x/n_z   \\
             + I_z\omega_y\omega_z - I_x\omega_y\omega_z
         \end{array}  \\
         \begin{array}{l}
             -\left[k_py_1 + k_d(n_z\omega_y-n_y\omega_z) + (n_x\omega_z-n_z\omega_y)\omega_z\right]I_y/n_z   \\
             + I_x\omega_x\omega_z - I_z\omega_x\omega_z
         \end{array}  \\
    \end{array}\right]
\end{array}.
\label{eq:tau_cmd}
\end{equation}

The thrust command $T_\mathrm{cmd}$ can be obtained from the desired vertical acceleration $\boldsymbol{a}_{z,\mathrm{des}}$ and (\ref{eq:F_dyn_equation}), yielding
\begin{equation}
T_\mathrm{cmd} = m(\boldsymbol{a}_{z,\mathrm{des}} + g) / \cos{\phi}\cos{\theta},
\label{eq:T_cmd}
\end{equation}
where $\theta$ and $\phi$ are pitch and roll angles of the quadrotor. 

Finally, substituting (\ref{eq:tau_cmd}) and (\ref{eq:T_cmd}) into (\ref{eq:reduced_control_allocation}), we can get the control input $\tilde{\boldsymbol{u}}$, namely thrust commands of the three remaining rotors
\begin{equation}
\tilde{\boldsymbol{u}} =  \tilde{\boldsymbol{G}}^{-1}\left[T_\mathrm{cmd},~\tau_{x,\mathrm{cmd}},~\tau_{y,\mathrm{cmd}}\right]^T.
\end{equation}

Note that this controller is modified from that in \cite{Sun2020} to avoid using motor speed measurements unobtainable for some platforms. As a consequence, this controller is less robust against model mismatch caused by, e.g., high-speed induced aerodynamic effects. If the rotor speeds are measurable, we refer readers to use the controller proposed in \cite{Sun2020} to improve the resilience against aerodynamic disturbances.

The proposed method also assumes that the motor failure is already known by the controller, which can be easily detected by methods such as monitoring the motor current~\cite{Moseler2000Application}, or using a Kalman filter~\cite{amoozgar2013experimental}.
\section{Onboard State Estimation}\label{sec:state_estimator}

\begin{figure}
    \centering
    \includegraphics[width=1.0\linewidth]{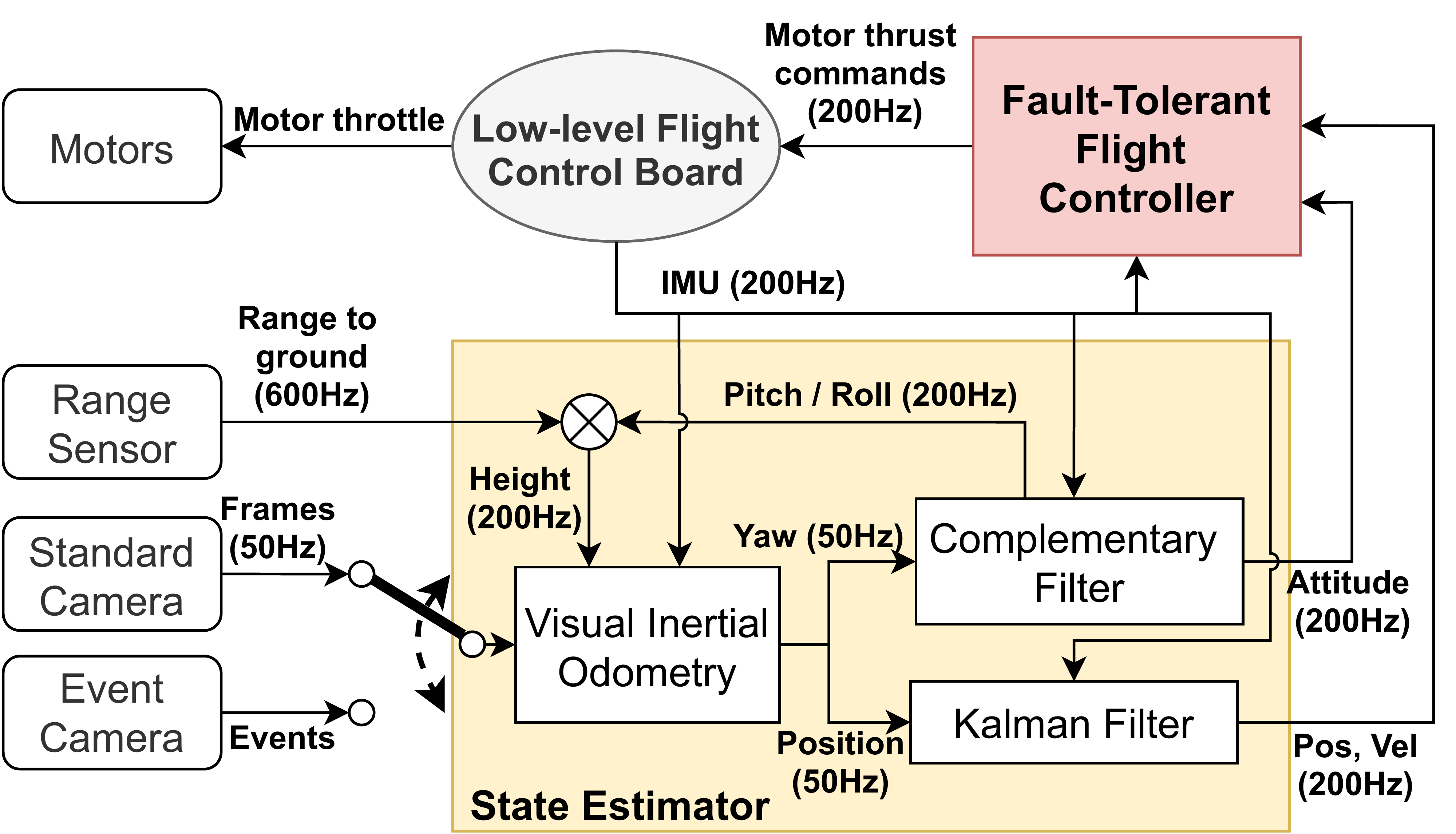}
    \caption{General diagram of the onboard state estimator and the fault-tolerant controller.}
    \label{fig:diagram_overall}
\end{figure}

To achieve autonomous flight with rotor failure, the state estimator provides orientation and position information using only onboard sensors, including an IMU, a downward looking range sensor, and a downward looking camera. The camera can be either a standard camera or an event camera. The block diagram of our system is given in Fig.~\ref{fig:diagram_overall}. A range-sensor-aided monocular VIO algorithm provides pose estimates of the camera at 50 Hz. These estimates are fused with the IMU from a low-level flight control board located at the center of gravity, using a Kalman filter and a complementary filter, to provide position, velocity, and orientation estimates at 200 Hz for the fault-tolerant flight controller. A rotation compensated complementary filter is proposed for orientation estimation instead of directly using the orientation from the VIO. This can improve the robustness of the algorithm in case of loss of feature tracks.

\subsection{Visual Inertial Odometry}
\label{sec:event_based_VIO}
\begin{figure}
    \centering
    \includegraphics[width=1.0\linewidth]{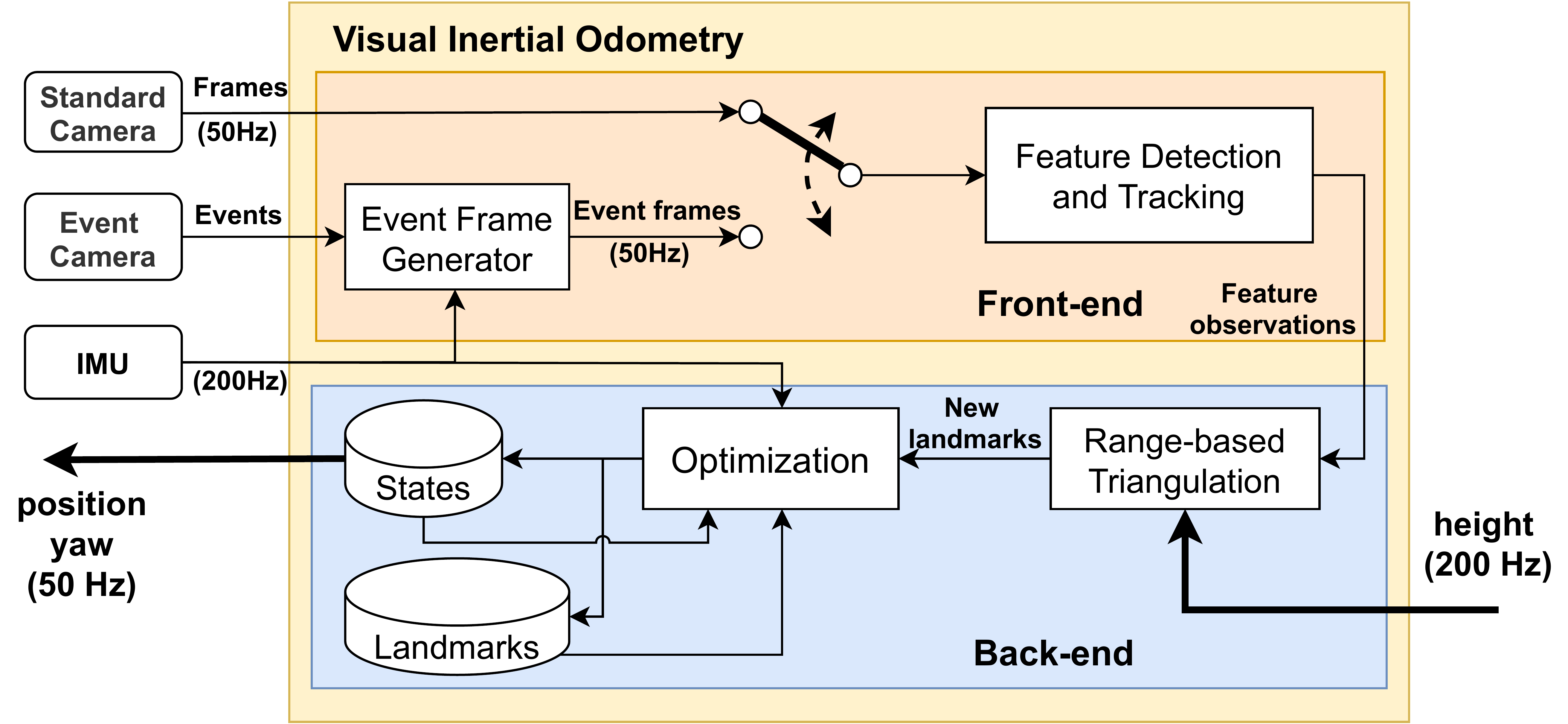}
    \caption{Diagram of the visual-inertial odometry (either event based or frame based).}
    \label{fig:diagram_eventvio}
\end{figure}

The proposed VIO provides position and yaw estimations. It can use either standard frames or events as visual inputs. A block diagram of the algorithm is given in Fig.~\ref{fig:diagram_eventvio}.

\subsubsection{Front-end}
If events are selected as visual input, we first of all generate synthetic event frames \cite{Rebecq2017}. Each event frame $I_k(\boldsymbol{x})$ is aggregated from events from a spatio-temporal window $W_k$: 
\begin{equation}
 I_k(\boldsymbol{x}) = \sum_{e_j\in W_k} p_j\delta (\boldsymbol{x} - \boldsymbol{x}{'}_j),
\end{equation}
where function $\delta(\cdot)$ is the Kronecker delta, $p_j$ is the polarity of a certain event represented by $e_j$, $\boldsymbol{x}'_j$ is the corrected event position considering the motion of the camera:
\begin{equation}
    \boldsymbol{x}{'}_j = \pi \left( T_{t_k, t_j} \left( Z(\boldsymbol{x}_i)\pi^{-1}(\boldsymbol{x}_i)\right) \right).
\end{equation}
where $\pi(\cdot)$ is the camera projection model, $T_{t_k, t_j}$ is the transformation of camera pose between $t_k$ and $t_j$, $Z(\boldsymbol{x}_i)$ is the scene depth approximated by the range sensor. 

As the rotation of the damaged quadrotor in this time window is more dominant than the linear motion, we use a pure rotation transformation to approximate $T_{t_k, t_{k-1}}$, which is generated by integrating angular rate measurements from the gyroscope. Then $T_{t_k, t_j}$ is approximated by a linear interpolation $T_{t_k, t_j} = T_{t_k, t_{k-1}} (t_j - t_{k-1})/(t_k - t_{k-1})$. 

Then we use a FAST feature detector~\cite{rosten2006machine} and Lucas-Kanade tracker~\cite{lucas1981iterative} to detect and track features. If a feature has been tracked over 3 consecutive frames, it is determined as persistent and is triangulated. The corresponding landmark is added to the map. These settings are the same as applied in~\cite{vidal2018ultimate} and~\cite{Rebecq2017}.

If standard frames are selected as visual input, we simply replace the synthetic event frames in the above procedure.

\subsubsection{Range-Based Landmark Triangulation}
\label{sec:landmark_triangulation}
Since the damaged quadrotor motion is mostly rotation, triangulating landmarks based on the disparity is inaccurate. For this reason, we use a downward range sensor to detect the range of the camera to the ground, where most features are detected. We also assume that all landmarks lie within the ground plane. For the $j$-th landmark whose position in the inertial coordinate frame is defined as $\boldsymbol{p}_j$ we have

\begin{equation}
\begin{split}
    \lambda \left[\begin{array}{c}
         u  \\
         v  \\
         1
    \end{array}\right]
    = \boldsymbol{K} \boldsymbol{R}_\mathrm{CI}
    \left(
    \boldsymbol{p}_j
    -
    \boldsymbol{p}_c
    \right),
    \\
     {p}_{c,z} - {p}_{j,z} = \hat{h}= r \cos\theta\cos\phi,\\
\end{split}
\label{eq:range_based_triangulation}
\end{equation}
where $u$, $v$ are observations of the landmark in the last (event) frame, $\boldsymbol{p}_c$ is the position of the camera in the inertial coordinate frame, $\boldsymbol{K}$ is the camera calibration matrix, $\boldsymbol{R}_\mathrm{CI}$ is the rotation matrix from camera frame to the inertial frame, $\hat{h}$ is the height estimate, $r$ is the range measurement from the range sensor.  

As $\boldsymbol{R}_\mathrm{CI}$ and $\boldsymbol{p}_c$ are obtained from the backend optimization, $\phi$ and $\theta$ are estimated from the complementary filter, we can subsequently solve $\boldsymbol{p}_j$ from (\ref{eq:range_based_triangulation}). 
The position estimates of triangulated landmarks are then used as initial guess for the nonlinear optimization problem in the backend.

\subsubsection{Back-end}
We use a keyframe-based fixed-lag smoother inspired by \cite{leutenegger2015keyframe} to estimate the pose of the camera. The optimization cost function is formulated as
\begin{equation}
\begin{split}
    J(\boldsymbol{X}) &= \sum^K_{k=1}\sum_{j\in J(k)}||\boldsymbol{e}^{j,k}_v||^2_{\boldsymbol{W}_v^{j,k}} + \sum^{K-1}_{k=1}||\boldsymbol{e}^{k}_i||^2_{\boldsymbol{W}_i^{k}} \\
    & + \sum^K_{k=1}\sum_{j\in J(k)}w_h^{j,k} ({p}_{c,z}-\hat{h}-{p}_{j,z})^2, 
    \label{eq:backend_optimization}
\end{split}
\end{equation}
where $k$ is the frame index, $K$ denotes number of frame in the sliding window, $j$ is the landmark index, $J(k)$ is the set containing all the visible landmarks from the frame $k$. The first and second term in (\ref{eq:backend_optimization}) represent reprojection error and inertial error respectively. The optimization variables are the states of the $K$ frames in the sliding window, which are represented by $\boldsymbol{X}$ and include position, velocity, orientation, and IMU biases. 

Differently from the optimization-based back-end in~\cite{leutenegger2015keyframe}, we add the third term in the cost function (\ref{eq:backend_optimization}), where $w_h^{j,k}$ is the weight. It forces the vertical differences between the position of the camera and the observed landmark to be equal to the height estimate $\hat{h}$ from the range sensor. By this means we add additional scale information from the range sensor while the IMU based scale information becomes less reliable in this fast rotational motion dominated task.

The optimization is run when a new (event) frame is generated. To improve computation efficiency, we do not perform marginalization and discard the states and measurements outside the sliding window. We use the Google Ceres Solver \cite{ceres} to solve this nonlinear least square problem.

\subsection{Rotation Corrected Complementary Filter}
The complementary filter is widely used to provide attitude estimates of a quadrotor. It has a major advantage that the pitch and roll are estimated only from IMU measurements, thus it is robust to failure of other sensors. The accelerometer measurement is expressed as:
\begin{equation}
    \boldsymbol{a}_\mathrm{IMU} = \boldsymbol{a}^B - \boldsymbol{g}^B + \boldsymbol{\omega}^B \times (\boldsymbol{\omega}^B \times \boldsymbol{d}_\mathrm{ba}^B) + \dot{\boldsymbol{\omega}}^B \times\boldsymbol{d}_\mathrm{ba}^B + \boldsymbol{w}_\mathrm{acc} + \boldsymbol{b}_\mathrm{acc}~,
    \label{eq:acc_measurement_model}
\end{equation}
where $\boldsymbol{a}^B$ is the acceleration of the center of gravity, $\boldsymbol{w}_\mathrm{acc}$ and $\boldsymbol{b}_\mathrm{acc}$ are the measurement noise and bias respectively. $\boldsymbol{d}_\mathrm{ba}^B$ is the displacement from the center of gravity to the accelerometer.

A standard complementary filter assumes that the accelerometer measures the negative gravitational vector expressed the body frame in a long-term period. In other words, it neglects other terms in (\ref{eq:acc_measurement_model}) except $\boldsymbol{g}^B$. This works well when a quadrotor spends significant periods of time in hover or slow forward flight~\cite{mahony2012multirotor}. However, when the quadrotor fast spins at a near constant body rate $\boldsymbol{\omega}^B$ and enter the relaxed-hovering condition \cite{Mueller2015}, $\boldsymbol{a}^B$ becomes a constant non-zero centripetal acceleration, and the displacement induced term $\boldsymbol{\omega}^B \times (\boldsymbol{\omega}^B \times \boldsymbol{d}_\mathrm{ba}^B)$ also becomes non-negligible. These two terms need to be considered in the filter, yielding the rotation corrected complementary filter.

\begin{figure}
    \centering
    \includegraphics[width=0.6\linewidth]{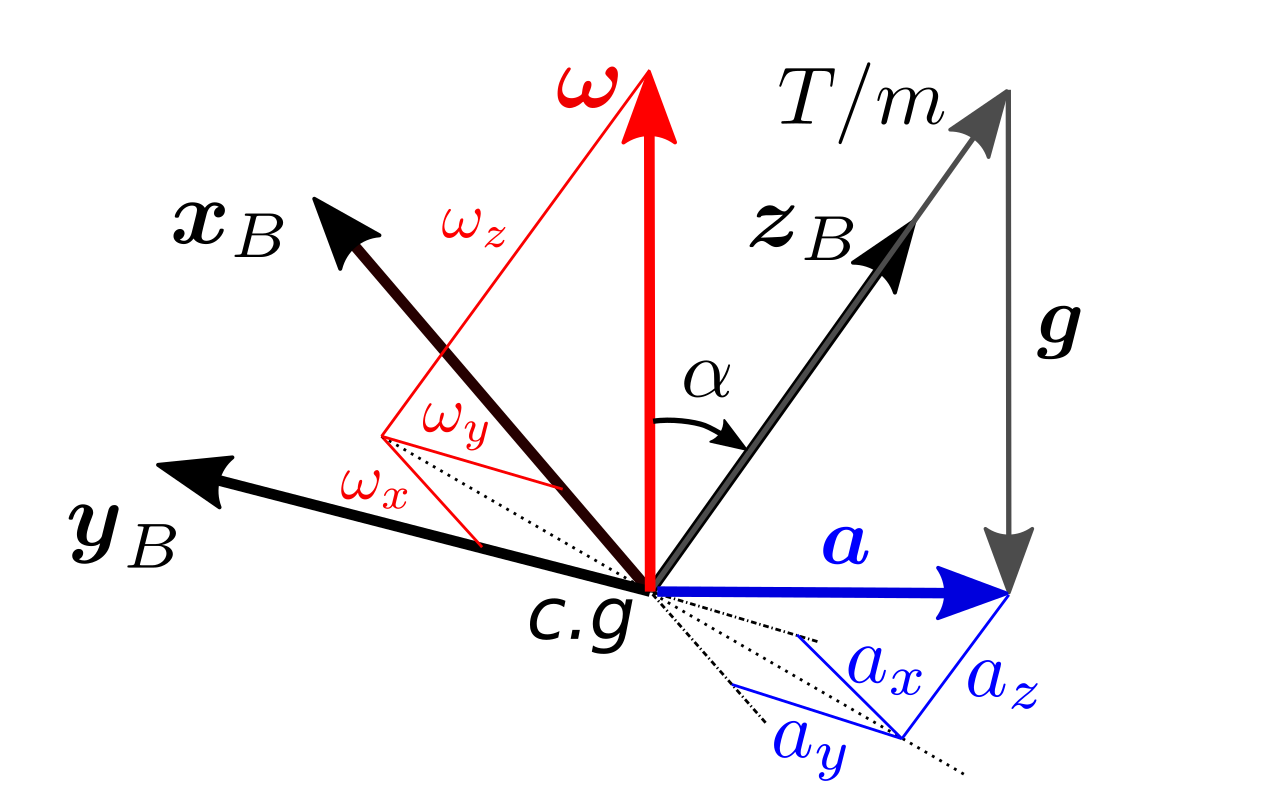}
    \caption{In the relaxed-hovering condition, the acceleration projection on the body frame ($\boldsymbol{a}^B$) is constant.}
    \label{fig:CF_vectors}
\end{figure}

We estimate acceleration $\boldsymbol{a}^B$ by assuming the quadrotor stays in the relaxed-hovering condition in a long-term period. According to Fig.~\ref{fig:CF_vectors}, we can obtain the estimated $\boldsymbol{a}^B$ as 
\begin{equation}
    \hat{\boldsymbol{a}}^B = \left[-\frac{\omega_x}{||\boldsymbol{\omega}||}g, ~-\frac{\omega_y}{||\boldsymbol{\omega}||}g,~g\tan{\alpha}\sin{\alpha}\right]^T,
    \label{eq:a_relaxed_hovering}
\end{equation}
where $\alpha = \arccos(\cos{\theta}\cos{\phi})$. Note that (\ref{eq:a_relaxed_hovering}) is valid only at relaxed-hovering condition where quadrotor fast spins. Therefore, we still assume a zero $\hat{\boldsymbol{a}}^B$ vector when $|\omega_z| < \bar{\omega}$, where $\bar{\omega}$ is a positive threshold. We use $\bar{\omega} = 10$~rad/s in our experiments. Knowing $\boldsymbol{a}^B$, we can subsequently estimate the displacement $\boldsymbol{d}_\mathrm{ba}$ from (\ref{eq:acc_measurement_model}) by a least-square estimator, using real flight data. 

\begin{figure}
    \centering
    \includegraphics[width=1.0\linewidth]{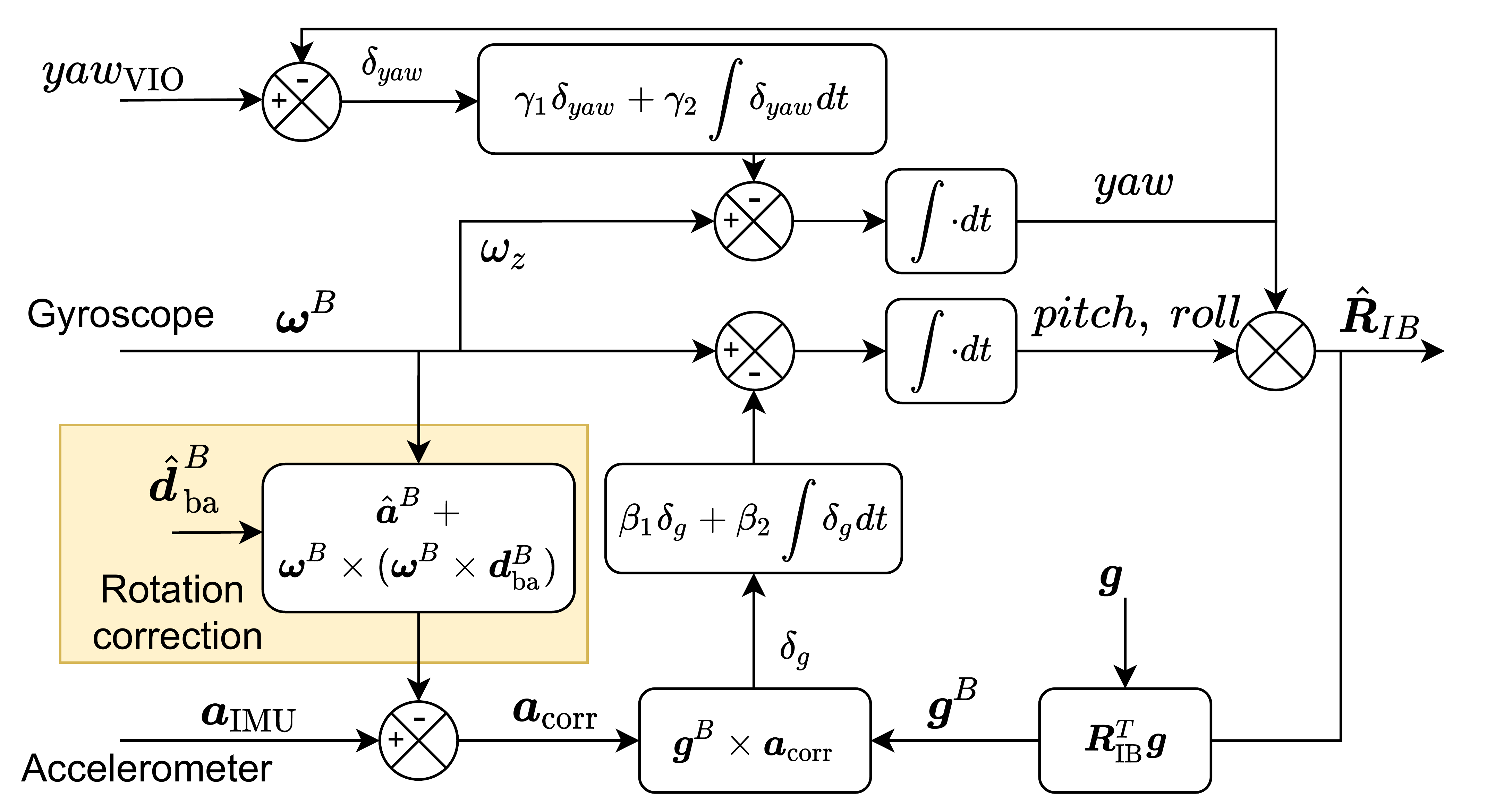}
    \caption{Diagram of the rotation corrected complementary filter.}
    \label{fig:diagram_complementary_filter}
\end{figure}

Fig.~\ref{fig:diagram_complementary_filter} presents the diagram of the proposed complementary filter, where the rotation compensation block is marked in yellow. Since there is no magnetometer used in our platform, we adopt the yaw angle from the VIO to fuse the gyroscope measurement and to obtain the yaw estimate. Finally, we can estimate the full attitude from this rotation corrected complementary filter.

\section{Experiments}\label{sec:experiments}
\subsection{Hardware Descriptions}
\begin{figure}[h!]
\begin{center}
\includegraphics[width=0.8\linewidth]{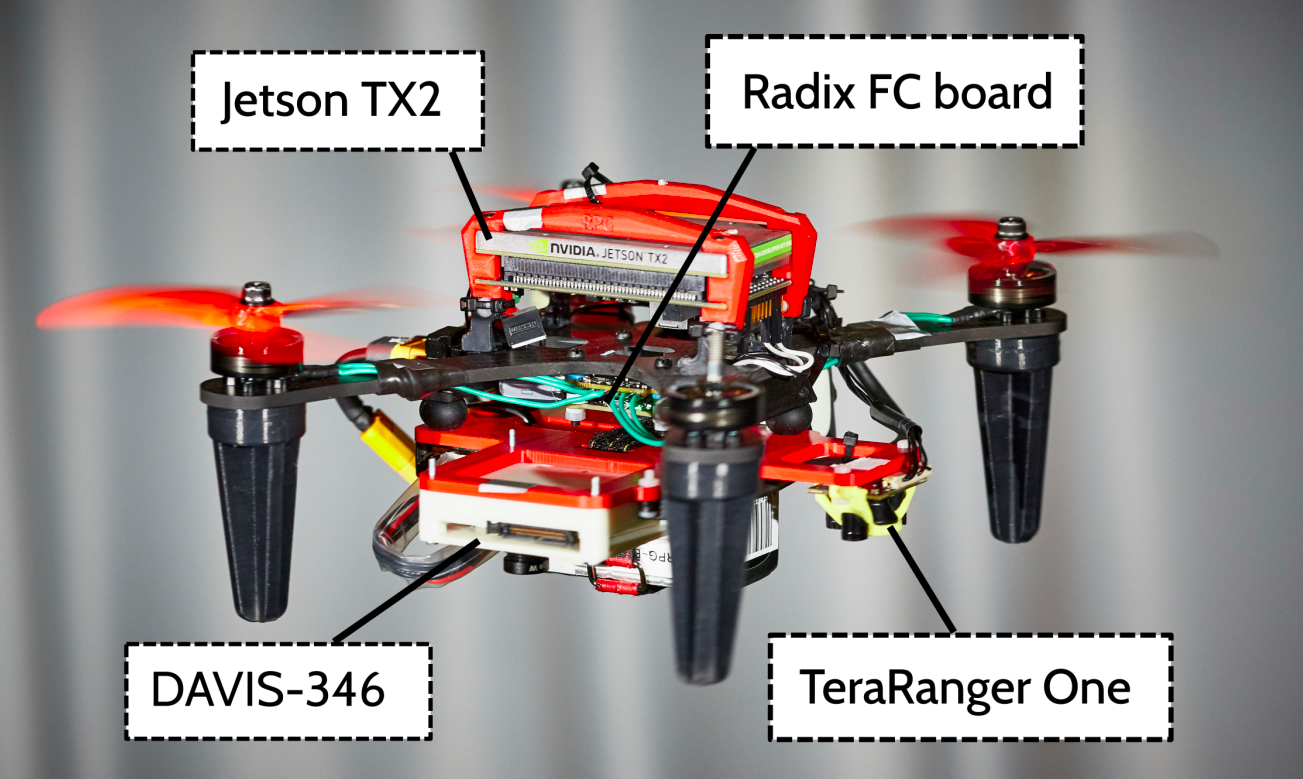}
\end{center}
  \caption{Photo of a quadrotor flying with three rotors, where an event camera is used for state estimation.}
\label{fig:drone}
\end{figure}
As Fig.~\ref{fig:drone} shows, the tested quadrotor is built with a carbon fiber frame and 3D printed parts. 
It is powered by four 2306-2400KV motors with 5-inch propellers. 
The state estimation and the control algorithm are run on a Nvidia Jetson TX2, which contains a quad-core 2.0~Hz CPU and a dual-core 2.5~Hz CPU running Ubuntu 18.04 and ROS \cite{quigley2009ros}. 
The motor thrust commands from the control algorithm are sent to motors through a Radix FC flight control board, which runs a self-built firmware that also sends the IMU measurements to the TX2 at 200~Hz. We used TeraRanger One, a LED time-of-flight range sensor, to measure the distance to the ground. Both standard and event cameras are facing downward. They both use a 110$^{\circ}$ field-of-view lens. For the event camera, we use an Inivation DAVIS-346 with a resolution of $346\times 240$ pixels. For the standard camera, we use a mvBlueFox-220w \cite{bluefox} with a resolution of $376 \times 240$ pixels, chosen intentionally to be close to that of the event camera to enable a fair comparison. It is worth noting that the maximum gain (12~dB) of the mvBlueFox-200w camera is used to minimize the required exposure time, which is found essential in reducing the motion blur in the standard frames.

\subsection{Validation of the Rotation Corrected Complementary Filter}
\begin{figure}
\begin{center}
\includegraphics[width=1.0\linewidth]{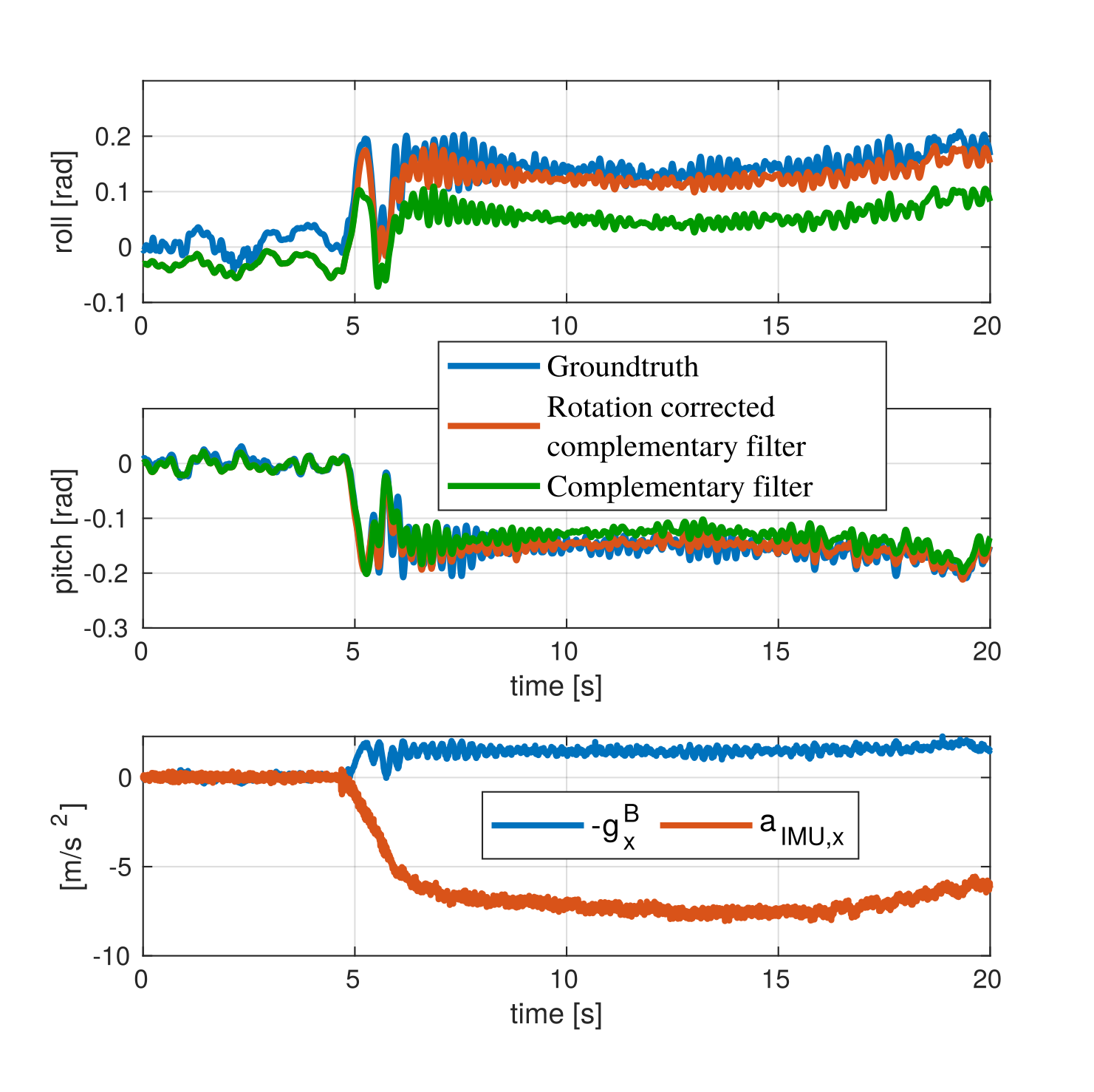}
\end{center}
  \caption{\textbf{Top two plots}: Pitch and roll angle estimates of the proposed rotation corrected complementary filter and a standard complementary filter, compared with the ground truth measured by the motion capture system. \textbf{Bottom plot}: comparison between the gravity projection on the body frame and the accelerometer measurement.}
\label{fig:CF_comparisons}
\end{figure}

We use the flight data to evaluate the performance of the proposed complementary filter, against the one without rotation motion considered. Fig.~\ref{fig:CF_comparisons} shows that the roll error of the standard complementary filter became large as the quadrotor started spinning at around 5~s. By contrast, the rotation corrected complementary filter could well estimate the pitch and roll angles despite the over 20 rad/s spinning rate. Bottom plot of Fig.~\ref{fig:CF_comparisons} explains the large error of the roll estimates. As standard complementary filter assumes that the accelerometer measures the negative gravitational vector, $\boldsymbol{g}^B$ should align with $\boldsymbol{a}_{\mathrm{IMU},x}$. However, this assumption becomes invalid when considerable centripetal acceleration appears, which leads to significant bias of the accelerometer measurement. 

\subsection{Closed-loop Flight Validation}


\begin{figure}
\centering
\begin{subfigure}{0.5\textwidth}
    \centering
     \includegraphics[width=0.8\linewidth]{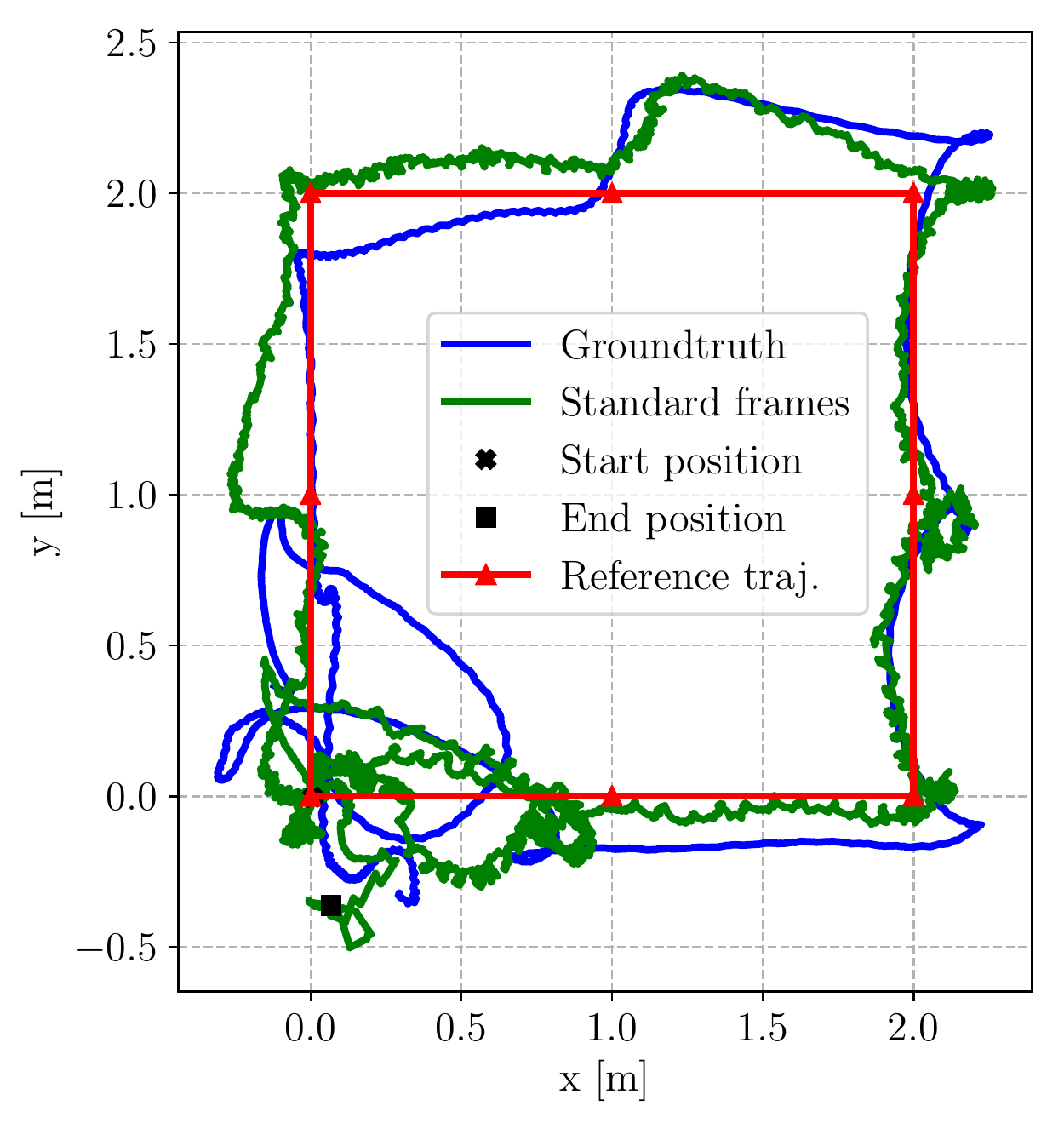}
    \caption{Closed-loop flight trajectory using \textbf{standard frames}.}
    \label{fig:closed_loop_frame}
\end{subfigure}
\newline
\begin{subfigure}{0.5\textwidth}
    \centering
    \includegraphics[width=0.8\linewidth]{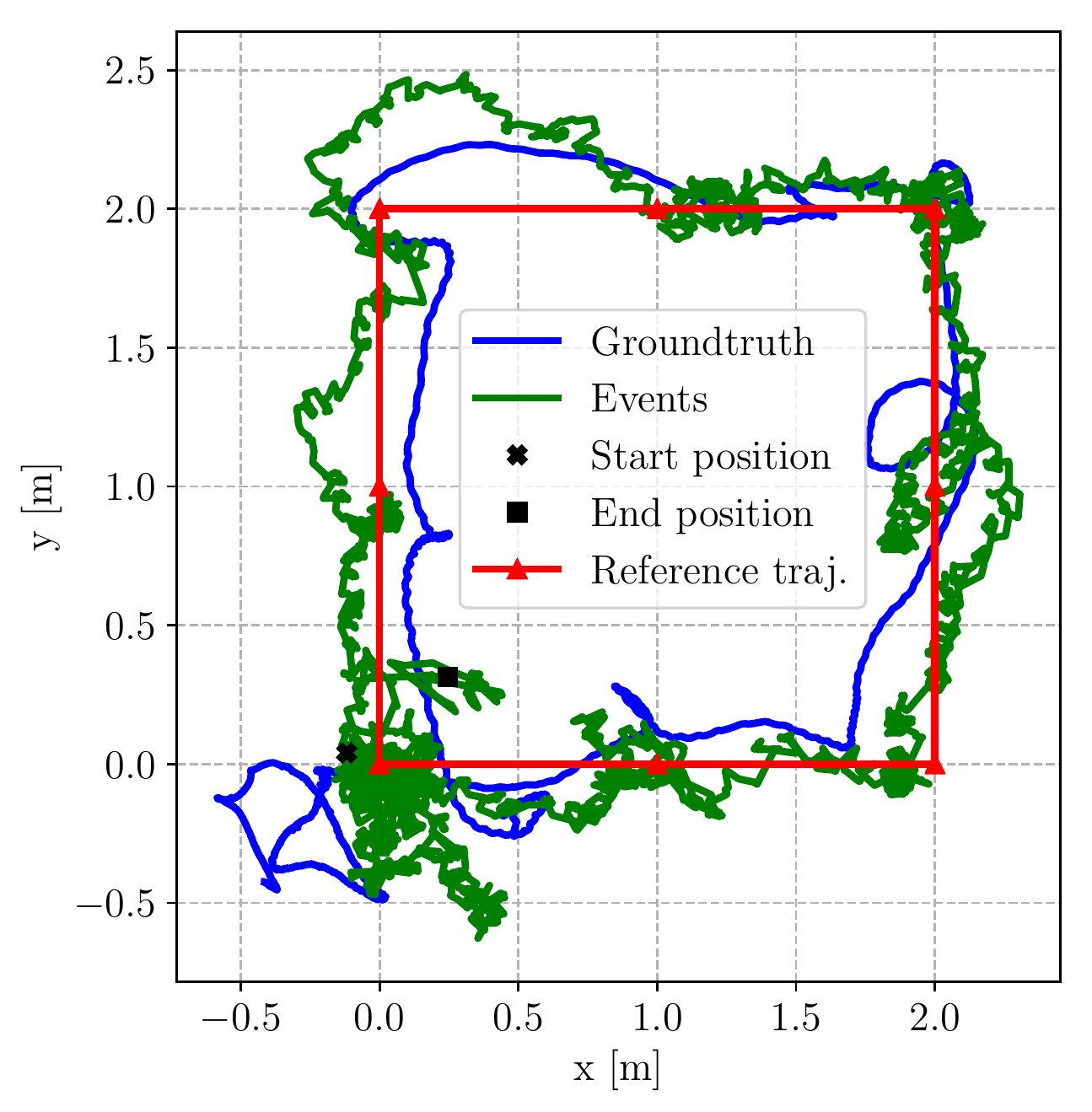}
    \caption{Closed-loop flight trajectory using \textbf{events}.}
    \label{fig:closed_loop_events}
\end{subfigure}
\caption{Top view of the closed-loop flight trajectories. Ground-truth from the motion capture system (blue), estimated trajectory (green), and the reference trajectory (red) including 9 setpoints.}
\end{figure}

We conducted setpoint tracking tasks in the closed-loop flight experiment to validate the entire algorithm, including the vision-based state estimator and the fault-tolerant flight controller. During the test, the quadrotor took off with four rotors. The VIO was initialized while the drone was in hovering. 
Then, we switched off one rotor and the fault-tolerant flight controller started controlling the quadrotor. 
As shown in Fig.~\ref{fig:closed_loop_frame}, nine setpoints formed into a square were given to the flight controller in steps of 5 seconds. The damaged quadrotor then flew a square trajectory by tracking these setpoints.

Two different tests were performed where standard frames and events were respectively used as visual input to the state estimator. Fig.~\ref{fig:closed_loop_frame} shows the closed-loop flight result using standard frames, where the estimated trajectory from the VIO in the inertial frame, the ground truth trajectory measured by the motion capture system, and the reference trajectory are presented. The difference between the real trajectory and the reference is caused by the position estimation error and the controller tracking error. Although the tracking performance is not perfect, it is sufficient for controlling a damaged quadrotor to a safe area for landing. Similarly, Fig.~\ref{fig:closed_loop_events} shows another test where events are used as visual input to the state estimator. Both tests were conducted in a well-illuminated environment (500 lux), which is a bright indoor office lighting condition~\cite{lux}.




\subsection{Comparison between Frames and Events}

\begin{table}[t]
    \caption{Position tracking accuracy (RMSE) with state estimators using standard frames or events in different light conditions. For standard frames, gains of the camera are set as 12~dB. Env lux: environment illuminance. Cam lux: illuminance at the camera lens. The left column shows photos of the test environment.}
    \centering
    \includegraphics[width=0.48\textwidth]{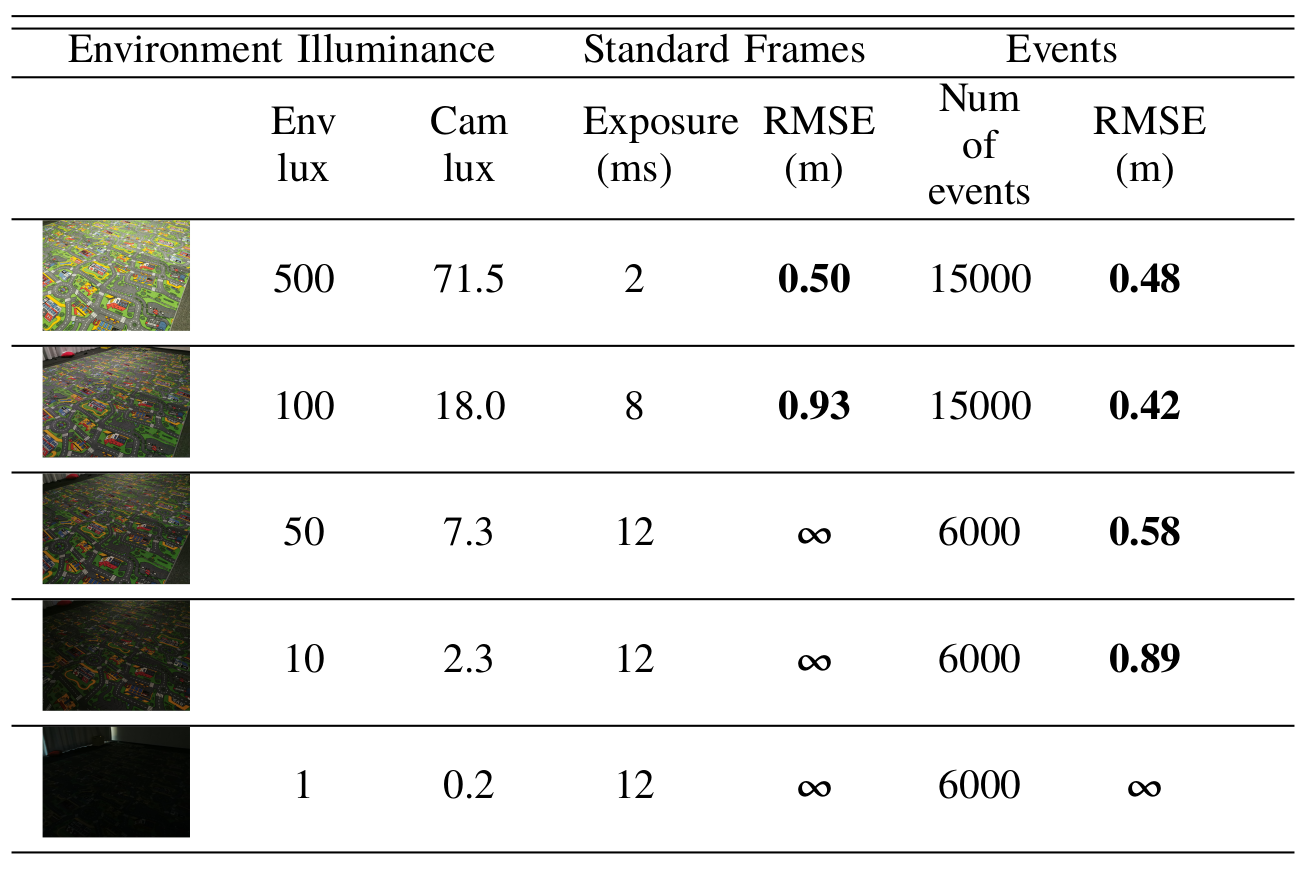}
\end{table}

In this section, we test the algorithm in different environment's lighting conditions, and compare between using frames and events as visual input. In these tests, we let the damaged quadrotor track a single setpoint (i.e., hovering). Then, we measure the root mean square error (RMSE) of the closed-loop position tracking to evaluate the performance of the entire algorithm. The exposure times of the standard camera are changed according to the environment brightness to capture frames with sufficient intensity for detecting and tracking features. For the event camera, we observe that the number of events generated in a fixed time window is smaller in a darker environment. Hence, we accordingly reduce the number of events needed to construct an event frame in low-light conditions.

Table~\ref{tab:illuminance_comparison} reports the position tracking RMSE when using standard frames or events in different lighting conditions. As can be observed, when the environment illuminance is around 500 lux, both  frames and events can accomplish the task with similar tracking error. However, with standard frames as visual input, the tracking error doubles as the illuminance  drops to 100 lux, and the damaged quadrotor crashes immediately when the illuminance gets lower than 100 lux. By contrast, with events as visual input, the damaged quadrotor can even fly when the illuminance is decreased to 10 lux. These comparisons clearly show the advantage of using an event camera for state estimation in low-light conditions.

\begin{figure}
\begin{center}
\includegraphics[width=0.8\linewidth]{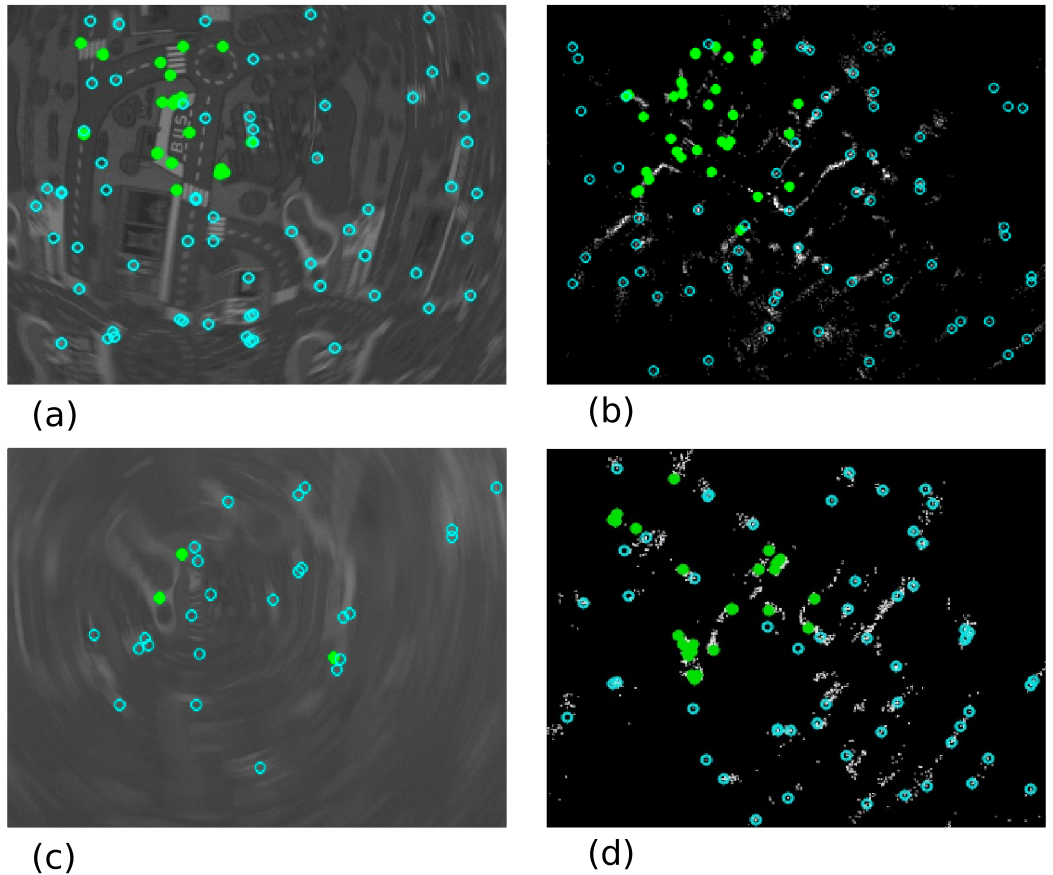}
\end{center}
  \caption{Standard and event frames, including features (light blue circle are detected features, green dots are persistent features), in a bright (500 lux) and a dark (50 lux) environment. (a) Standard frame in the bright environment with 2~ms exposure time. (b) Event frame in the bright environment. (c) Standard frame in the dark environment with 12~ms exposure time. (d) Event frame in the dark environment.
  }
\label{fig:frame_comparison}
\end{figure}

Fig.~\ref{fig:frame_comparison} presents the standard frames and the event frames in a bright (500 lux) and a dark (50 lux) indoor environment, respectively. When the environment illuminance is 50 lux, a relatively long exposure time (12~ms) of the standard camera is required. Hence, the standard frames experience significant motion blur (Fig.~\ref{fig:frame_comparison}c) owning to the quadrotor fast rotational motion caused by the motor failure. Although more than 20 features are detected from this blurry image, few of them are successfully tracked (i.e., persistent features) and added to the map, causing failure of the standard frame-based VIO. By contrast, the event frames are sharp enough for feature detection and tracking in both bright and dark conditions.

\subsection{Outdoor Flight Test}
\begin{figure}
\begin{center}
\includegraphics[width=1.0\linewidth]{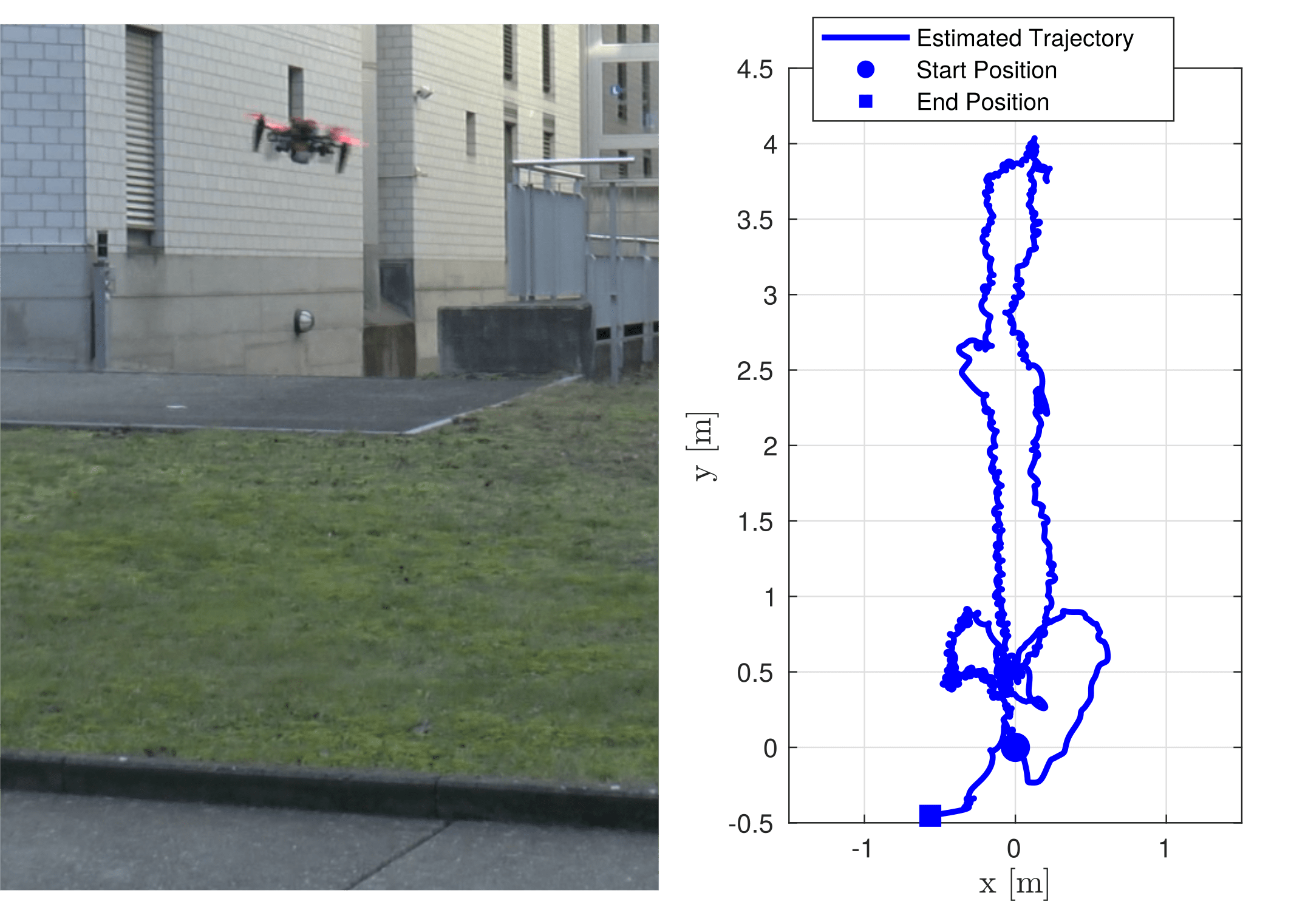}
\end{center}
  \caption{Outdoor flight to validate the proposed fault-tolerant flight solution. \textbf{Left}: snapshot of the qudarotor flying with only three propeller. \textbf{Right}: top view of the forward-backward flight trajectory given by the state estimator.}
\label{fig:outdoor}
\end{figure}
Finally, we validate the proposed algorithm in an outdoor environment with natural light and textures. Fig.~\ref{fig:outdoor} shows the snapshot of the flying quadrotor, and the top view of the flight trajectory from the state estimator. The quadrotor was commanded to conduct forward flight for 4 meters, and then return to the origin. The entire flight can be found in the supplementary video. In this flight, the environment illuminance was 2000 lux. According to Table~\ref{tab:illuminance_comparison}, both standard frames and events are reliable in such bright conditions. Hence, a standard camera was used in this flight.

\section{Conclusions}\label{sec:conclusions}

In this work, to the best of our knowledge we achieved the first autonomous flight of a quadrotor despite loss of a single rotor, using only onboard sensors. A new state estimation pipeline was proposed, including a rotation corrected complementary filter and a VIO algorithm aided by a range sensor. Despite the fast spinning motion of the damaged quadrotor, we demonstrated that the proposed method can provide reliable state estimates to perform hovering flights and setpoint-tracking tasks with only onboard sensors.

Comparisons were made between different visual inputs to the proposed algorithm: standard frames and events. In a well-illuminated environment, we demonstrated that the algorithms using both forms of visual input can provide sufficiently accurate state estimates. However, in a relatively low-light environment with illuminance lower than 100 lux, the standard frames were affected by significant motion blur due to the fast rotation and the long exposure time required. By contrast, the event-based algorithm could stand closed-loop tests in a much darker environment (10 lux).

Finally, we conducted outdoor flight tests to validate the proposed method in realistic conditions, with natural light and texture. 

We believe that the this work can significantly improve quadrotor flight safety in both GPS denied or degraded environments.

\section*{ACKNOWLEDGMENT}
We appreciate valuable discussions and suggestions from Zichao Zhang, Henri Rebecq, and Philipp Föhn. We also appreciate the help from Thomas Längle, Manuel Sutter, Roberto Tazzari, Tobias Büchli, Yunlong Song, Daniel Gehrig, and Elia Kaufmann.


\bibliographystyle{ieeetr}
\bibliography{IEEEabrv,mendeley}

\end{document}